\documentclass{article}

    \PassOptionsToPackage{numbers, compress}{natbib}

    \usepackage[preprint]{neurips_2024}

\usepackage[utf8]{inputenc} 
\usepackage[T1]{fontenc}    
\usepackage{hyperref}       
\usepackage{url}            
\usepackage{booktabs}       
\usepackage{amsfonts}       
\usepackage{nicefrac}       
\usepackage{microtype}      
\usepackage{xcolor}         
\usepackage{wrapfig}
\usepackage{graphicx} 
\usepackage{xcolor,colortbl}
\usepackage{booktabs,multirow}

\usepackage{subcaption}

\newcommand{\yz}[1]{}
\renewcommand{\yz}[1]{\textcolor{red}{\textbf{Yun}: #1}}

\newcommand{\ls}[1]{}
\renewcommand{\ls}[1]{\textcolor{magenta}{\textbf{LS}: #1}}

\title{Accelerating Inference of Retrieval-Augmented Generation via Sparse Context Selection}

\author{
  Yun Zhu$^1$, Jia-Chen Gu$^3$, Caitlin Sikora$^2$, Ho Ko$^2$, Yinxiao Liu$^1$, Chu-Cheng Lin$^2$,  \\
  {\bf Lei Shu$^1$, Liangchen Luo$^1$, Lei Meng$^2$, Bang Liu$^4$, Jindong Chen$^1$} \\
  $^1$Google DeepMind \\
  $^2$Google \\
  $^3$University of California, Los Angeles \\
  $^4$Université de Montréal \& Mila \\
  \texttt{\{yunzhu,csikora,hoko,canoee,kitsing,leishu,luolc,leimeng,jdchen\}@google.com} \\
  \texttt{gujc@ucla.edu}, \texttt{bang.liu@umontreal.ca}
}

\begin{document}

\maketitle

\begin{abstract}
Large language models (LLMs) augmented with retrieval exhibit robust performance and extensive versatility by incorporating external contexts. 
However, the input length grows linearly in the number of retrieved documents, causing a dramatic increase in latency.
In this paper, we propose a novel paradigm named Sparse RAG, which seeks to cut computation costs through sparsity.
Specifically, Sparse RAG encodes retrieved documents in parallel, which eliminates latency introduced by long-range attention of retrieved documents.
Then, LLMs selectively decode the output by only attending to highly relevant caches auto-regressively, which are chosen via prompting LLMs with special control tokens. 
It is notable that Sparse RAG combines the assessment of each individual document and the generation of the response into a single process.
The designed sparse mechanism in a RAG system can facilitate the reduction of the number of documents loaded during decoding for accelerating the inference of the RAG system.
Additionally, filtering out undesirable contexts enhances the model's focus on relevant context, inherently improving its generation quality.
Evaluation results of two datasets show that Sparse RAG can strike an optimal balance between generation quality and computational efficiency, demonstrating its generalizability across both short- and long-form generation tasks.

\end{abstract}

\section{Introduction}

Large language models (LLMs) have attracted increasing attention and exhibited impressive abilities to understand instructions and generate fluent outputs in natural language~\citep{gpt3, gpt3.5, llama, team2023gemini}. 
Nevertheless, LLMs inevitably manifest hallucinations~\citep{DBLP:journals/csur/JiLFYSXIBMF23} due to their struggle with factual errors and inability to secure the accuracy of generated text solely by the parametric knowledge they encapsulate~\cite{hallucination-survey, factuality-evaluation}. 
Feeding the source of truth to LLMs in the format of retrieved context segments~\citep{reid2024gemini} alleviates this problem. The technique is widely known as Retrieval-Augmented Generation (RAG)~\cite{rag,rag-survey,rag-pretrain}.

Although the RAG framework is empirically shown to be effective, it can be expensive to scale up. This is because it requires prepending relevant documents retrieved from an external knowledge corpus to the queries~\cite{rag-pretrain}. As a result, the input length grows linearly in the number of documents, causing a dramatic increase in latency when using a standard Transformer whose the latency scales quadratically with the input length.
Some prior works such as Fusion-in-Decoder (FiD)~\citep{izacard2020leveraging} and Parallel Context Windows (PCW)~\citep{ratner2022parallel} have proposed to alleviate this issue. Yet these methods fail to strike an optimal balance between generation quality and computational efficiency.
FiD was originally designed for the encoder-decoder architecture, and thus is not compatible with currently prevalent decoder-only architectures without significant changes. While
PCW can be applied to decoder-only LLMs, it only speeds up the model pre-filling and still incurs high latency since the whole context window cache is 
still being attended to when decoding each token.
Moreover, the heavy reliance of generation on the retrieved knowledge raises significant concerns about the model's behavior and performance in scenarios where retrieval may fail or return inaccurate results~\cite{irrelevant-ctx}. 
A typical approach for mitigating this issue is to rely on an external classifier to rank or filter the documents before prepending them to the input~\cite{yan2024corrective}, but this process requires extra model calls which adds new complexity to inference.

\begin{figure}[t]
  \centering
  \includegraphics[width=0.95\textwidth]{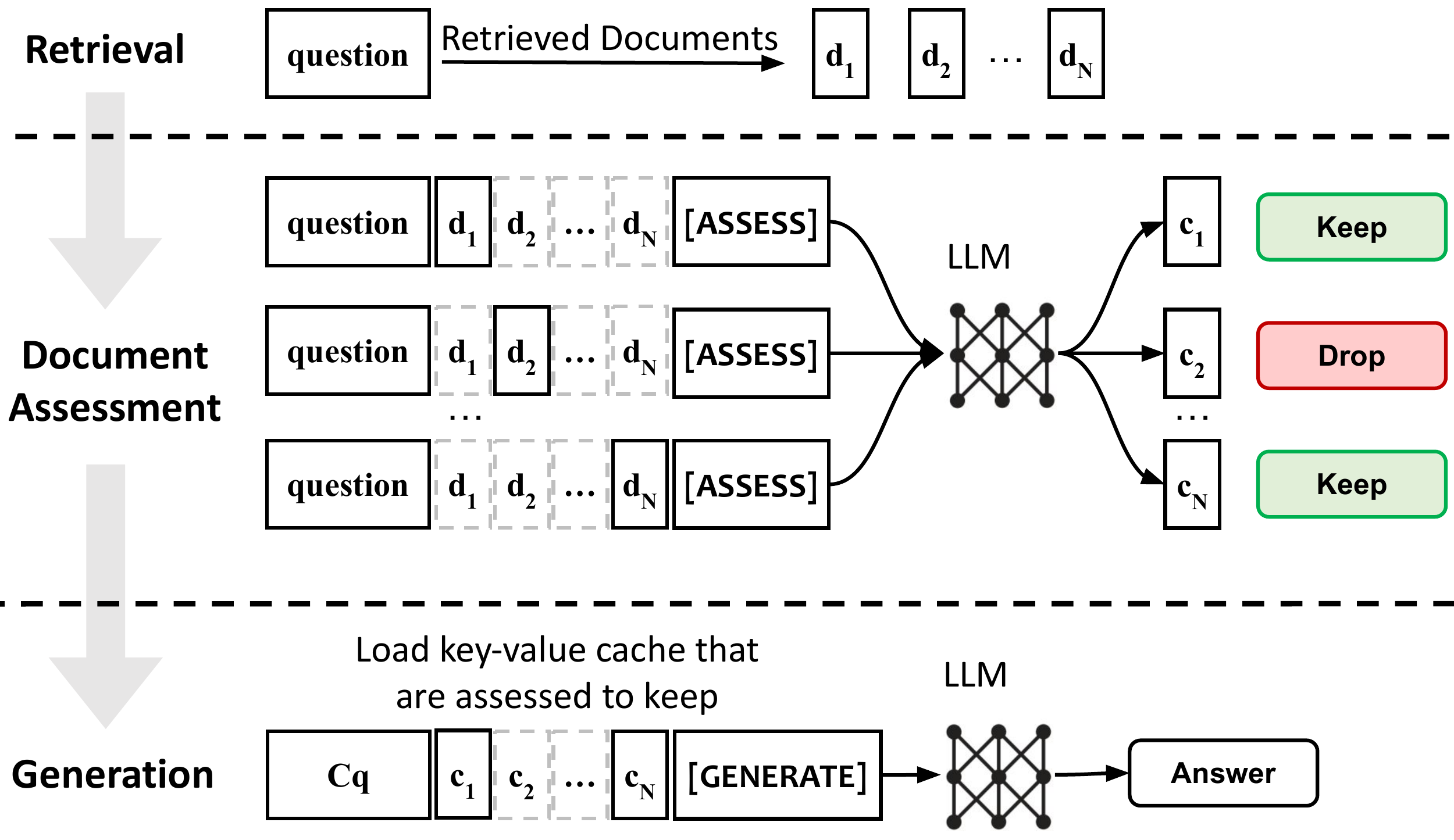}
  \caption{
   An overview of Sparse RAG at inference. 
   Each of the retrieved documents will be assessed by LLMs to decide whether to keep or drop by estimating a relevance score.
   Then, load the documents that are considered to keep for generation.
  } 
   \vspace{-3mm}
  \label{architecture}
\end{figure}

In light of the issues above, we propose a novel paradigm named Sparse RAG.
It generally operationalizes by massive pre-filling where the key-value cache was generated through a single forward pass of the input tokens and selective decoding where the output is generated by attending to highly relevant tokens auto-regressively. 
Previous works where the length of the retrieved contexts during pre-filling are equal to that during decoding are called \emph{dense}-RAG in this paper.
Different from that, Sparse RAG allows the pre-filled context to be significantly larger than the decoding context, where partial low-quality context is dynamically dropped according to their relevance to the input query. 
Furthermore, Sparse RAG combines the assessment of each individual context and the generation of the response into a single process. 
Special control tokens are used to prompt LLMs to assess each retrieved context for picking high-quality ones. 
Then only the key-value caches of those picked contexts are loaded into LLMs for decoding with another control tokens. 

The design of such a sparse mechanism in RAG system brings two unique advantages. 
Firstly, by reducing the number of key-value cache loads during the decoding process, LLMs can achieve lower latency as they are typically constrained by memory usage.
Besides, filtering out undesirable contexts enhances the model's focus on relevant context, inherently improving its generation quality.
To demonstrate the effectiveness and efficiency of the proposed method, we evaluate on two datasets of PopQA~\cite{popqa}, and Biography~\cite{factscore}.
Experimental results show that Sparse RAG can achieve the similar or better performance while keep much better latency compared with standard dense-RAG or PCW-RAG, demonstrating its generalizability across both short- and long-form generation tasks.

\section{Related Work}

\paragraph{Retrieval-Augmented Generation} 

RAG is a family of techniques for generating output structures while using retrieved nearest-neighbor structures as a reference. It typically involves two stages: retrieval and generation. 
Retrieval finds most similar contexts based on BM25 or learned embedding, where the context could be represented as token embedding~\citep{khandelwal2019generalization, yogatama2021adaptive}, dense embedding~\citep{de2021mention} or raw text~\citep{guu2020retrieval, izacard2020leveraging, lewis2020retrieval}. 
Once those contexts are retrieved, different architectures are leveraged to incorporate them into the model. 
Popular approaches includes concatenation~\citep{izacard2020leveraging, lewis2020retrieval} and cross-attention~\citep{borgeaud2022improving, lewis2020pre}.
In recent years, the LLM architecture evolves towards a single decoder only model with significant larger sizes. To this end, concatenation of raw text~\citep{lewis2020retrieval} is becoming popular for being simple and practical.  
Many advanced approaches have been developed on top of it in recent years.
Self-RAG~\cite{self-rag} is proposed to selectively retrieve knowledge and introduce a critic model to decide whether to retrieve.
\citet{ralm-nli} designed an NLI model to identify the irrelevant context and improve robustness.
SAIL \cite{sail} is tuned on instructions to insert retrieved documents before instructions.
Toolformer~\cite{toolformer} is pre-trained for calling APIs such as Wikipedia.
\citet{active-rag} actively anticipate future content and decide when and what to retrieve in long-form generation. 
CRAG~\citep{yan2024corrective} makes the attempt to explore and design corrective strategies for RAG to improve its robustness of generation. Specifically, an external T5 model is trained and used to determine the usefulness of the retrieved context. 
Generally, these approaches target on exploiting retrieval as a useful tool to augment generation or whether retrieval is necessary.


\vspace{-2mm}
\paragraph{Efficiency in RAG}
The efficiency of LLM inference is a widely explored research area, where different categories of approaches have been studied. Some works are towards architecture level acceleration such as efficient attention~\citep{shazeer2019fast}, Mixture of Expert~\citep{fedus2022switch}, Transformer-alternative architectures~\citep{gu2023mamba}, etc. While some other works are algorithm-level acceleration like quantization~\citep{lin2023awq} or speculative decoding~\citep{leviathan2023fast}. 
Nevertheless, most previous works are not RAG specific. 
Recently, RAG Cache~\citep{jin2024ragcache} was proposed as a multilevel dynamic caching system tailored for RAG from the system perspective. 
Another stream of the work try to process each document individually so that no cross document attention computation is required, such as FiD~\citep{izacard2020leveraging} or PCW~\citep{ratner2022parallel}. 
FiD encoded each retrieved passage independently from other passages and decoded by attending over the concatenation of the resulting representations of all the retrieved passages. 
PCW carved a long context into chunks (``windows''), restrict the attention mechanism to apply only within each window, and re-use the positional embeddings across the windows.



Comparison with previous works that are the most relevant to our work is illustrated in Table~\ref{tab:my_label}.
This work aims to strike an optimal balance between generation quality and computational efficiency.
It is notable that the extra classifier in CRAG requires maintaining extra model with more complex serving infrastructure. When there is $N$ contexts retrieved. There is totally $N+1$ model runs. 
Our work also relies an classification process to refine the retrieved documents. However, our approach relies on the ``internal'' classification process that is aligned with the generation process. The total model run in our case is $1$.


\begin{table}[t]
    \centering
    \caption{Comparisons with existing RAG-related works.}
    \resizebox{0.85\linewidth}{!}{
    \begin{tabular}{ccccc}
      \toprule
         Approach&Corrective&No extra model&Prefill efficiency&Decode efficiency \\
      \midrule
         RAG~\citep{lewis2020retrieval} & No& Yes& No & No  \\
         Corrective RAG~\cite{yan2024corrective} & Yes & No & No & No \\
         PCW RAG~\citep{ratner2022parallel} & No & Yes & Yes & No \\
         Sparse RAG &Yes& Yes& Yes& Yes \\
      \bottomrule
    \end{tabular}
    }
    \label{tab:my_label}
    \vspace{-3mm}
\end{table}
\section{Sparse RAG}

Sparse RAG is designed for the decoder-only model architecture, which is the default case for most popular LLMs. 
Figure~\ref{architecture} presents an overview of Sparse RAG at inference, which designs the document assessment strategies to improve the robustness of generation.
The key hypothesis of our approach is that the RAG task and per context assessment is a similar tasks and the model could do both tasks in one shot.
We found simple and effective approaches to enable them. 


\subsection{Training Process}

Our work is based on the assumption that a certain amount of RAG training data is accessible, which allows us to effectively tailor and adapt existing LLMs to our specific needs. 
In the training phase, we integrate an additional Per Context Assessment (PCA) task into the training mixture. 
This PCA task is included alongside the primary RAG task to enhance the model's ability to assess and respond accurately within various contexts. 
By incorporating the PCA task, we aim to improve the overall performance and contextual understanding of the LLMs, ensuring they are more adept at handling a diverse range of scenarios presented during their use.

\paragraph{Data Augmentation with LLMs}

For typical RAG data, one question-answer pair can be mapped to multiple retrieved contexts using either BM25 or existing standalone retriever. However, there are cases where no golden label regarding the quality of every retrieved context is available.

To collect these missing labels, we leveraged two off-the-shelf LLMs \citep{palm2, team2023gemini}–PALM2 and Gemini–to get the assessment for each context. We observe empirically that a second round of prompting for critique, especially using a different model from the initial round, ensures the best quality labels. We provide our prompts in Table~\ref{tab:critique_prompt} in the Appendix. We compare different model combinations for labeling to human ground truth labels in Section~\ref{human-labels}.

\paragraph{Multitasking Data Format}

The LLM is trained on a mixture of two types of tasks, ``Rate'' assessment and answer generation. Specifically, we format the inputs and outputs of the two task types as
\begin{itemize}
    \item Assessment: \{Question\}\{Context\}\{Control\_Assessment\}\{Rate\}
    \item Generation:  \{Question\}\{Context\_1\}...\{Context\_N\}\{Control\_Generation\}\{Answer\}
\end{itemize}
where {Rate} and {Answer} are targets for the generative tasks while everything before them are inputs. Control\_Assessment and Control\_Generation are fixed tokens to ensure the LLM can differentiate the two tasks.  

\paragraph{Parallel Contexts}
Since each context is rated independently in PCA task, we introduce this independence in the primary RAG task as well so two tasks could reuse KV cache during inference time. 
Thus for the generation task, we enforce no cross attention between different two contexts following the Parallel Context Windows~\citep{ratner2022parallel} to process the input for training. 
Specifically, we modify two things in the standard LM training process. First, we 
change the attention masks to be block-wise, and restrict that {context\_i} and {context\_j} will not attend to each other. {Question}, {Context\_i} and {Control\_Generation} still follow the default attention mechanism, i.e. in the causal attention case, the latter could attend to the previous ones.   
Second, we adjust the position encoding to ``mimic'' the situation that {Context\_i} is not visible to {context\_j} that each context will have its own incremental positional ids. However, the position id of {Control\_Generation} is designed to back to normal, where its position considers all the length of contexts.

\subsection{Inference Process}
Given the question and retrieved contexts, Sparse RAG handles the assessment task and generation task in one single pass. 

\paragraph{Per Context Assessment}
Similar to the training process, when pre-filling the KV cache, each retrieved context is treated independently. Then these KV cache will be used to score the input, which is the concatenation of the token of {Control\_Assessment} and the token of ``Good''. The score is the  probability of this tokens as the next tokens.  

\paragraph{Generation}
The generation uses a filtered KV cache, where only $K$ out of $N$ KV cache will finally be loaded. We use a simple threshold based filtering approach: we drop the context when its score is less than $sigma$. Once those caches are loaded, {Control\_Generation} will be used to prompt the model to generate the answers.




\section{Evaluation of Per Context Assessment}
\label{human-labels}

\subsection{New Benchmark:  Natural Question Per Context Assessment}

We isolated a subset of 50 questions, each with 10 retrieved contexts, from the Natural Questions dataset. We assigned 3 raters to each question-context pair from a pool of 7 raters and provided the instructions in Section~\ref{rater}.

We aggregated responses for all 3 raters for each context, selecting the majority decision 0 or 1 for each context. We found that raters unanimously agreed on 351 out of 500 context, with ~30\% of the documents considered relevant. For questions where raters were not unanimously decided, a specialist rater was assigned to investigate more carefully and set the best label to correct mistakes of the other raters. 
This resulted in 6 additional documents considered relevant out of the entire dataset, boosting the portion of relevant documents to 31\% and slightly increasing alignment with the auto-rater approaches (average F-score increase of 1.4\% across auto-rater methods using these corrections as the ground truth).

\subsection{LLM rater comparisons}

We tested several different LLM-based automatic labeling methods–different combinations of models and prompts–for creating training data for the classifier in Sparse RAG. We compared several of these auto-rater approaches by creating a ground-truth relevance dataset using human labeling. 
The auto-rater comparison using the revised human labels as the ground-truth is shown in Table~\ref{tab:autorater_comparison}. We find that combining two different models in two rounds–initial prompting and critique–provides the labels that are most closely aligned with the human labels. We hypothesize that the different representations learned by two different models are able to capture the most nuance in the input sequences, leading to better relevance judgements. We also observe that Gemini Ultra appears slightly less effective at critiquing model outputs than PALM2 XL.

\begin{table}[t]
    \centering
    \caption{Auto-rater comparison to ground truth.}
    \resizebox{0.78\linewidth}{!}{
        \begin{tabular}{c c c c c}
          \toprule
          \multicolumn{2}{c}{Auto-labeling method} & \multicolumn{1}{c}{Average F1}  & \multicolumn{1}{c}{F1 Label 0} & \multicolumn{1}{c}{F1 Label 1} \\
          Rater model & Critic model & & \\
          \midrule
          PALM2 XL & n/a & 0.729 &	0.765 &	0.694 \\
          PALM2 XL & PALM2 XL & 0.781 &	0.820 &	0.741 \\
          Gemini Ultra & n/a & 0.761 &	0.807 &	0.716 \\
          Gemini Ultra & Gemini Ultra & 0.704 &	0.747 &	0.660 \\
          PALM2 XL & Gemini Ultra & 0.728 &	0.776 &	0.680 \\
          Gemini Ultra & PALM2 XL & \textbf{0.821} &	\textbf{0.861} &	\textbf{0.781} \\
          \bottomrule
        \end{tabular}
    }
    \label{tab:autorater_comparison}
    \vspace{-2mm}
\end{table}

\section{Evaluation of Sparse RAG}

\subsection{Benchmarks and Metrics}

\paragraph{PopQA} is a large-scale open-domain question answering (QA) dataset, consisting of 14k entity-centric QA pairs. Each question is created by converting a knowledge tuple retrieved from Wikidata using a template. We follows the setup from ~\citep{yan2024corrective} and use Contriver~\citep{contriever} to retrieve the related contexts. 
Since PopQA is lacking the per-context assessment label, we adopted the ``Gemini + PALM2'' combination to get training labels. We split the dataset into training, validation and test set with 8:1:1 ratio. 
Since the answer is usually short, we report \textbf{Exact Match (EM)} and \textbf{F1} scores.

\paragraph{QMSum}~\citep{zhong2021qmsum} is a human-annotated benchmark for query-based multi-domain meeting summarization task, which consists of 1,808 query-summary pairs over 232 meetings in multiple domains. To adopt it in the RAG domain, we divide each conversation into different contexts where the average context contains 300 words, and filter the data point where the  conversation containing ground truth label is split into multiple contexts.
Note that this dataset has human labeled per-context assessment that we could leverage during training.
Eventually we get $250$ training data, $70$ validation and $77$ test set. The target of this data is longer and we report \textbf{RougeLSum} and \textbf{F1} scores. 

\subsection{Baselines}

\paragraph{RAG} We evaluated the standard RAG~\citep{rag} where an LM generates output given the query prepended with all the top ranked documents using the same retriever as
in our system.


\paragraph{PCW RAG} We applied the Parallel Context Windows~\citep{ratner2022parallel} to the RAG process, where no cross attention is applied between documents.  

\paragraph{Corrective RAG} It is the same with RAG except that an external T5-XXL classifier is trained using heuristic labels~\citep{yan2024corrective}. This classifier is used to process all the document and decide the rank. Note to facilitate the fair comparison, we did not adopt the "web search" feature of this paper. 

\subsection{Experimental Configuration}

The base LLMs used in the paper Gemini~\citep{team2023gemini}. 
Although our approach could be applied to different training stage of the model,  we apply LoRA tuning~\cite{hu2021lora} to enforce alignment on top of the foundation LLMs due to its low resource requirement and wide usage. Note that the same LoRA tunning is applied to Sparse RAG and all baselines. In all our experiment, we apply LoRA in self-attention and use the default rank as 4. By default, we use the XXS size which could be running on-device.  

During training, we use $64$ Tensor Processing Units (TPU) V3 chips for PopQA while use $128$ Units for . 
The batch size is $64$. 
We use the Adafactor optimizer~\citep{shazeer2018adafactor} with a learning rate of $0.003$. The training dropout rate is $0.05$. 
We leverage the metrics of validation set to pick the best checkpoint.
During inference, the temperature is set to $0.5$. Unless specifically noted, we use sampling decoding with sample number $1$ for our experiments.

\subsection{Inference Setup and Metrics}

Evaluation of Sparse RAG was conducted on a Samsung S21 Ultra, utilizing the device's CPU to assess real-world performance on a relatively mid-tier smartphone compared to the latest flagship models. Inference configuration consisted of fixed token lengths for queries, contexts and generated responses. This setup allows for evaluating the system's efficiency and effectiveness under resource constraints typical of mobile devices, providing insights into its practical applicability for on-device question-answering tasks.

Specifically, the overall inference process considers two stages.
\begin{itemize}
    \item Prefill stage:
     For baseline RAG Model, we measure the total time taken to process all input tokens (question and all contexts).
     For PCW RAG and Sparse RAG models: We take advantage of these models' ability to cache question processing. We first measure the time to process the question alone. Then, we measure the time to process different contexts with the pre-processed kv-cached question. We use \textbf{Encoding Speed (ES)} (tokens per second (t/s)) to measure the efficiency of prefill stage.  
    \item Decoding stage:
       To assess decoding speed comprehensively, we generate output sequences using the same question but varying the amount of relevant context data considered, ranging from the top 1 most relevant document to the top $K$. For each context size, we produce output sequences of different lengths. This systematic approach allows us to evaluate the impact of both context size and response length on decoding speed. We use \textbf{Decoding Speed (DS)} (tokens per second (t/s)) to measure the efficiency of decoding stage. 
\end{itemize}







\subsection{Main Results}

We report the results with both short form answer (PopQA) and long form answer (QMSum) in Table~\ref{tab:main}, where both quality and latency metrics are shared. ``K'' is the number of chosen memories.

Notably, our proposed approach achieves the best quality while being the most efficient during inference compared to other ``Dense RAG'' approaches. 
It can be seen that Sparse RAG shares the same pre-filling efficiency with PCW, due to the parallel context encoding. However Sparse RAG can achieve significant better decoding efficiency than standard RAG and PCW RAG. 
In detail, out of 20 retrieved contexts, the Sparse RAG has average of $7.84$ context for PopQA and $4.45$ context for QMSum. This leads to almost \textbf{double} or even \textbf{triple} decoding speed.
Meanwhile, Sparse RAG achieves higher quality metrics than the dense counterparts, which shows that Sparse RAG effectively filtered noisy and irrelevant contexts. 

On the other hand, it is interesting to compare Corrective-RAG(CRAG) with Sparse RAG on PopQA task since Corrective-RAG trained their own external classifier with T5 XXL based on PopQA data. Results suggested that our approach, albeit using an ``in-place'' classifier, is outperforming the CRAG with external classifier. Note the encoding and decoding speed of CRAG is not comparable because it involves multiple model running from the classifier.

\begin{table}[t]
\centering
\caption{Quality \& Efficiency Tradeoff.  For both short form and long form generation tasks, Sparse RAG achieves both higher quality and efficiency compared to ``dense'' RAG approaches.}
\begin{tabular}{l|c|cccc|cccc}
\toprule
  & \multirow{2}{*}{ES} & \multicolumn{4}{c|}{PopQA}
  & \multicolumn{4}{c}{QMSum}\\
   &  &
  EM &
  F1 &
  K &
  DS &
  F1 &
  RougeLSum &
  K&
  DS \\

  \midrule
RAG~\citep{lewis2020retrieval}   &56.28     & 65.43  & 69.99 &20 & 
6.65  &21.43&18.20&20
&6.64\\
PCW-RAG\citep{ratner2022parallel}  & \textbf{147.58}      & 65.04  & 69.54 &20
& 6.65 &20.18&16.95&20&6.64 \\
CRAG~\citep{yan2024corrective}     &-   & 66.52  & 70.99   & 8.90 & - &-&-&-&- \\
Sparse RAG       &\textbf{147.58}   & \textbf{67.17}  & \textbf{71.16} & \textbf{7.84}& 
\textbf{12.28} &\textbf{23.96}&\textbf{20.10}&\textbf{4.45}& \textbf{16.05} \\
\bottomrule
\end{tabular}

\label{tab:main}
\end{table}

\subsection{Analysis}

\subsubsection{Impact of Confidence Threshold}

To show how the metrics changes with different thresholds, the performance of Sparse RAG corresponding to different thresholds is illustrated in Table~\ref{tab:threshold}.
As the threshold gradually increases, the system will filter out more contexts, which would always reduce the latency during inference.
In terms of generation quality, it can be seen that the performance of Sparse RAG was significantly improved as thresholds increased at the beginning, which shows the effectiveness of filtering out irrelevant contexts. 
Then, the performance was stable and dropped slightly. 
The reason might be that some truly relevant contexts are accidentally filtered out.

\begin{table}[ht]
\centering
\caption{Sampling a few confidence threshold on PopQA and QMSum. Higher confidence threshold means filtering more memories. Efficiency wise, filtering more always helps; While in terms of quality.}
\begin{tabular}{c|cccc|cccc}
\toprule
  & \multicolumn{4}{c|}{PopQA}
  & \multicolumn{4}{c}{QMSum}\\
  Threshold &
  EM &
  F1 &
  K &
  DS &
  F1 &
  RougeLSum &
  K&
  DS \\
  \midrule
0.05        & 66.95 &	70.97 &  9.75 & 10.61  &22.85 &	19.49& 7.92& 11.92 \\
0.1        & 66.84	& 70.66 &  8.72 & 11.70 &23.78&	19.98 &6.68& 12.89\\
0.15      & \textbf{67.17}	& \textbf{71.16} &  7.84 & 12.28  & 23.43	& 19.66 &5.77& 13.01 \\
0.2     & 66.77  &70.54	 &  7.13 & 12.88 & 23.2 &	19.79  &5.05& 14.54 \\
0.25        & 65.75&	69.64  &  6.56 & 13.00 & \textbf{23.96}	&\textbf{20.1} &4.45& 16.05 \\
0.3        & 63.86  & 68.2 &  \textbf{5.98} & \textbf{13.08} & 23.84 &	19.99 &\textbf{3.93}& \textbf{16.38} \\
\bottomrule
\end{tabular}
\label{tab:threshold}
\end{table}

\subsubsection{Silver Labels vs LLM Labels}

In Corrective RAG, the T5 model was trained with silver labels that comes from the title matching~\citep{yan2024corrective}. We collect the same silver labels to replace the LLM labels and train the model with this new dataset. The comparison result is shown in Table~\ref{tab:heurestic}.

From these results, we can observe that the quality of the labels generated by LLMs is higher than that of the silver labels from \citet{yan2024corrective}. 
We hypothesize that the superior quality of the LLM-generated labels is attributable to our methodology, which involved a two-round process of soliciting responses from two different LLMs. 
By engaging two distinct models, we likely enhanced the robustness and accuracy of the labels through a form of cross-validation, thereby mitigating potential biases or errors that might arise from relying on a single LLM.

\begin{table}[ht]
\centering
\caption{Comparing silver labels with LLM labels on PopQA.}
\begin{tabular}{lllll}
\toprule
  Approach &
  EM &
  F1 &
  K&
  DS \\\midrule

Sparse RAG     & \textbf{67.17}  & \textbf{71.16} & \textbf{7.84}& 
\textbf{12.28} \\
w/ silver label     & {66.97}  & {71.05} & {8.26}& 
{11.99}  \\
\bottomrule
\end{tabular}
\label{tab:heurestic}
\end{table}

\subsubsection{Ablation of Prefill Documents}
One question one may have is that the better performance may rely on the ``massive'' prefilling but in practical situation, we can retrieve smaller amount of context which have better quality. 
To this end, we compare different numbers of the prefill document: 10 vs 20 for PopQA, and show the results in Table~\ref{tab:num_prefill}.
Even in the setting lower prefilled case, the previous conclusion of Sparse RAG remains. It achieves better quality with higher speed of decoding. 
It makes sense that the gap between dense RAG and sparse RAG is relative smaller at 10 prefill documents.  This is because the headroom is smaller as the top 10 documents naturally has less to filter and make it less ``sparse''. Yet our approach still achieves best trade-off, showing the capability to generalize with different numbers of input documents.

\begin{table}[ht]
\centering
\caption{Ablation on Different number of prefill documents for PopQA.}
\begin{tabular}{lcccccc}
\toprule
  Approach &
  Prefill Documents &
  EM &
  F1 &
  K &
  ES &
  DS \\\midrule
RAG &10 & 64.66	& 68.67 &  10& 80.74 &10.31  \\
PCW RAG &10 &63.9	&68.58 &10& \textbf{147.48} & 10.31 \\
Sparse RAG&10        & \textbf{65.86} & 	\textbf{70.2} & \textbf{7.79} &\textbf{147.48}&\textbf{12.33}   \\
\midrule
RAG&20       & 65.43  & 69.99   &20&56.28&  6.65\\
PCW RAG&20       & 65.04  & 69.95  &20&147.58 &  6.65 \\
Sparse RAG&20       & 67.17  & 71.16   &   7.84 &147.58 & 12.28 \\
\bottomrule
\end{tabular}
\label{tab:num_prefill}
\end{table}

\subsubsection{Inference Efficiency Ablations}

Focusing on the decoding stage, we delve into the performance differences when varying the number of retrieved memories (i.e., top-K documents) and the length of the generated response. This analysis provides valuable insights into the computational overhead associated with incorporating additional context and generating more extensive outputs, ultimately shedding light on the practical trade-offs for real-time question-answering on resource-constrained devices.

\begin{figure*}[ht]
    \begin{subfigure}{0.5\linewidth}
        \includegraphics[width=\linewidth]{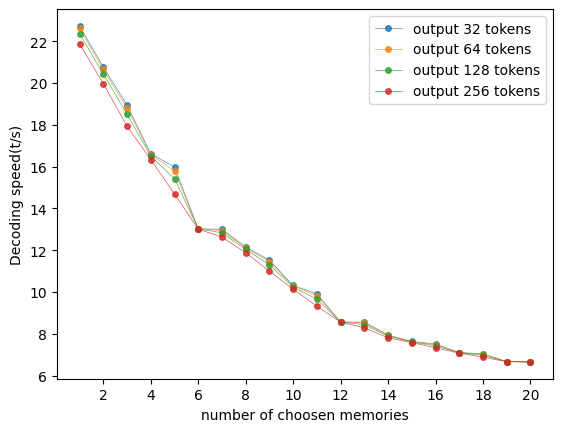}
        \caption{Decoding speed with different number of memories}
        \label{fig:a}
    \end{subfigure}
    \hfill
    \begin{subfigure}{0.5\linewidth}
        \centering
        \includegraphics[width=\linewidth]{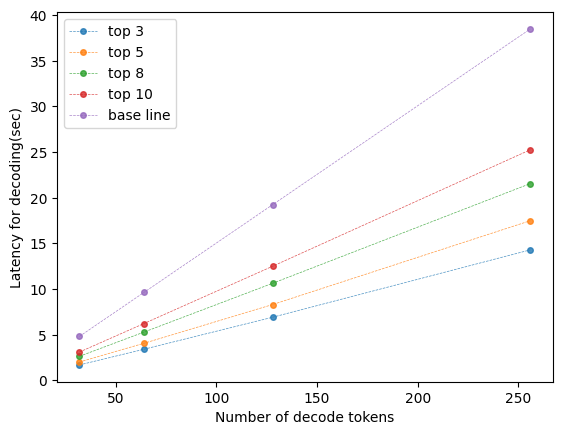}
     \caption{E2E Latency for decoding different number of tokens}
     \label{fig:b}
    \end{subfigure}
    \caption{Inference Efficiency Comparison}
    \label{fig:latency}
\end{figure*}

As illustrated in Fig~\ref{fig:b}, we observed that the baseline method requires over 50\%(varies from 52\% to over 55\% depending on number of output tokens) more time to generate outputs of varying lengths compared to the Sparse RAG approach. This significant disparity in generation speed is primarily attributed to the increased computational burden incurred by the baseline method when considering a larger number of memories. As shown in Fig~\ref{fig:a}, this heightened demand for computational resources results in a notable slowdown in terms of tokens per second (t/s). This observation underscores the efficiency advantages offered by Sparse RAG, especially in scenarios where a substantial amount of context needs to be considered during the decoding process.

\subsubsection{Ablation on Foundation Model Size}

We applied Sparse RAG to different sizes of LLMs, specifically testing it on Gemini XS and Gemini XXS. 
The results of these experiments are presented in Table~\ref{tab:size}. 
The findings demonstrate that Sparse RAG is compatible with various foundation models, effectively adapting to different model sizes. 
Notably, with a reduced amount of decoding caches, Sparse RAG is capable of achieving the highest quality results. 
This indicates that Sparse RAG maintains its efficiency and effectiveness across different foundation models, making it a versatile approach for various LLM configurations.

\begin{table}[ht]
\centering
\caption{Ablation on Different LLM model sizes.}
\begin{tabular}{lcccc}
\toprule
  Approach &
  Model Size &
  EM &
  F1 &
  K \\\midrule
RAG &XS & 66.52	& 70.87 &  20  \\
PCW RAG &XS &65.75	&70.37 &20 \\
Sparse RAG&XS        & \textbf{68.26} & 	\textbf{72.26} & \textbf{6.27}   \\
\midrule
RAG&XXS       & 65.43  & 69.99   &20\\
PCW RAG&XXS       & 65.04  & 69.95  &20 \\
Sparse RAG&XXS       & 67.17  & 71.16   &   7.84  \\
\bottomrule
\end{tabular}
\label{tab:size}
\end{table}

\subsubsection{Using Golden Context Label during inference}


Since QMSum provides golden per-context labels, we leverage these labels during inference to evaluate the upper bound performance under the condition of perfect per-context assessment. 
By utilizing the highest quality labels available, we aim to determine the best possible outcomes that our model can achieve. 
The results of this experiment, which highlight the performance ceiling under ideal labeling conditions, are presented in Table~\ref{tab:golden}. 
This approach allows us to understand the maximum potential of our model when supplied with optimal input data, thereby offering insights into its ultimate capabilities.

\begin{table}[ht]
\centering
\caption{Trying golden labels on QMSum.}
\begin{tabular}{lllll}
\toprule
  Approach &
  F1 &
  RougeLSum &
  K&
  DS \\\midrule

Sparse RAG     & {23.96}	&{20.1} &4.45& 16.05  \\
+ golden label     & \textbf{26.76}	& \textbf{21.93} &  \textbf{1.13} & \textbf{21.16}  \\
\bottomrule
\end{tabular}
\label{tab:golden}
\end{table}

\section{Conclusion}
This paper presents Sparse RAG to address the challenges of increased input length and latency. 
Through a novel approach of massive pre-filling and selective decoding, Sparse-RAG efficiently manages the key-value cache of retrieved documents, allowing the LLMs to focus on highly relevant tokens. 
This selective attention mechanism not only reduces the computational burden during inference but also enhances the generation quality by filtering out irrelevant contexts. 
The evaluation on two datasets validates Sparse RAG's ability to achieve a balanced trade-off between high-quality generation and computational efficiency, proving its versatility and effectiveness for both short- and long-form content generation tasks. 
This innovative paradigm showcases the potential for improving LLM performance in various applications by optimizing context management and inference processes.
\section*{Limitation and Future Work}

Sparse RAG requires additional tuning on top of the LLMs, which is not a plug-and-use approach. 
Under cases where per-context assessment label is missing, the prompts used for LLM rating may need to be adjusted for different use cases.
These adjustments ensure that the model can still function effectively despite the lack of specific context labels.
Furthermore, future research will explore Sparse-RAG in multimodal contexts, investigating how Sparse-RAG can handle and integrate information from multiple types of data to improve its performance and applicability across diverse scenarios.

\section{Acknowledge}
We would like to thanks Yu-hui Chen, Yuqi Li, Qifei Wang, Zonglin Li, Zhong Meng, Alex Chen, Jingbin Wang, Abhanshu Sharma, Alec Go, Jesper Andersen, David Racz, DeLesley Hutchins, Luke Sernau, David Munday, and Ewa Dominowska for their suggestions. 
\newpage
\bibliography{custom}

\begin{thebibliography}{40}
\expandafter\ifx\csname natexlab\endcsname\relax\def\natexlab#1{#1}\fi

\bibitem[{Anil et~al.(2023)Anil, Dai, Firat, Johnson, Lepikhin, Passos,
  Shakeri, Taropa, Bailey, Chen, Chu, Clark, Shafey, Huang, Meier{-}Hellstern,
  Mishra, Moreira, Omernick, Robinson, Ruder et~al.}]{palm2}
Rohan Anil, Andrew~M. Dai, Orhan Firat, Melvin Johnson, Dmitry Lepikhin,
  Alexandre Passos, Siamak Shakeri, Emanuel Taropa, Paige Bailey, Zhifeng Chen,
  Eric Chu, Jonathan~H. Clark, Laurent~El Shafey, Yanping Huang, Kathy
  Meier{-}Hellstern, Gaurav Mishra, Erica Moreira, Mark Omernick, Kevin
  Robinson, Sebastian Ruder, et~al. 2023.
\newblock \href {https://doi.org/10.48550/ARXIV.2305.10403} {{PaLM} 2 technical
  report}.
\newblock \emph{CoRR}, abs/2305.10403.

\bibitem[{Asai et~al.(2023)Asai, Wu, Wang, Sil, and Hajishirzi}]{self-rag}
Akari Asai, Zeqiu Wu, Yizhong Wang, Avirup Sil, and Hannaneh Hajishirzi. 2023.
\newblock \href {https://doi.org/10.48550/ARXIV.2310.11511} {Self-rag: Learning
  to retrieve, generate, and critique through self-reflection}.
\newblock \emph{CoRR}, abs/2310.11511.

\bibitem[{Borgeaud et~al.(2022)Borgeaud, Mensch, Hoffmann, Cai, Rutherford,
  Millican, Van Den~Driessche, Lespiau, Damoc, Clark
  et~al.}]{borgeaud2022improving}
Sebastian Borgeaud, Arthur Mensch, Jordan Hoffmann, Trevor Cai, Eliza
  Rutherford, Katie Millican, George~Bm Van Den~Driessche, Jean-Baptiste
  Lespiau, Bogdan Damoc, Aidan Clark, et~al. 2022.
\newblock Improving language models by retrieving from trillions of tokens.
\newblock In \emph{International conference on machine learning}, pages
  2206--2240. PMLR.

\bibitem[{Brown et~al.(2020)Brown, Mann, Ryder et~al.}]{gpt3}
Tom~B Brown, Benjamin Mann, Nick Ryder, et~al. 2020.
\newblock Language models are few-shot learners.
\newblock In \emph{Advances in neural information processing systems}, pages
  1877--1901.

\bibitem[{De~Jong et~al.(2021)De~Jong, Zemlyanskiy, FitzGerald, Sha, and
  Cohen}]{de2021mention}
Michiel De~Jong, Yury Zemlyanskiy, Nicholas FitzGerald, Fei Sha, and William
  Cohen. 2021.
\newblock Mention memory: incorporating textual knowledge into transformers
  through entity mention attention.
\newblock \emph{arXiv preprint arXiv:2110.06176}.

\bibitem[{Fedus et~al.(2022)Fedus, Zoph, and Shazeer}]{fedus2022switch}
William Fedus, Barret Zoph, and Noam Shazeer. 2022.
\newblock Switch transformers: Scaling to trillion parameter models with simple
  and efficient sparsity.
\newblock \emph{Journal of Machine Learning Research}, 23(120):1--39.

\bibitem[{Gu and Dao(2023)}]{gu2023mamba}
Albert Gu and Tri Dao. 2023.
\newblock Mamba: Linear-time sequence modeling with selective state spaces.
\newblock \emph{arXiv preprint arXiv:2312.00752}.

\bibitem[{Guu et~al.(2020{\natexlab{a}})Guu, Lee, Tung, Pasupat, and
  Chang}]{rag-pretrain}
Kelvin Guu, Kenton Lee, Zora Tung, Panupong Pasupat, and Ming{-}Wei Chang.
  2020{\natexlab{a}}.
\newblock \href {http://proceedings.mlr.press/v119/guu20a.html} {Retrieval
  augmented language model pre-training}.
\newblock In \emph{Proceedings of the 37th International Conference on Machine
  Learning, {ICML} 2020, 13-18 July 2020, Virtual Event}, volume 119 of
  \emph{Proceedings of Machine Learning Research}, pages 3929--3938. {PMLR}.

\bibitem[{Guu et~al.(2020{\natexlab{b}})Guu, Lee, Tung, Pasupat, and
  Chang}]{guu2020retrieval}
Kelvin Guu, Kenton Lee, Zora Tung, Panupong Pasupat, and Mingwei Chang.
  2020{\natexlab{b}}.
\newblock Retrieval augmented language model pre-training.
\newblock In \emph{International conference on machine learning}, pages
  3929--3938. PMLR.

\bibitem[{Hu et~al.(2021)Hu, Shen, Wallis, Allen-Zhu, Li, Wang, Wang, and
  Chen}]{hu2021lora}
Edward~J Hu, Yelong Shen, Phillip Wallis, Zeyuan Allen-Zhu, Yuanzhi Li, Shean
  Wang, Lu~Wang, and Weizhu Chen. 2021.
\newblock Lora: Low-rank adaptation of large language models.
\newblock \emph{arXiv preprint arXiv:2106.09685}.

\bibitem[{Izacard et~al.(2022)Izacard, Caron, Hosseini, Riedel, Bojanowski,
  Joulin, and Grave}]{contriever}
Gautier Izacard, Mathilde Caron, Lucas Hosseini, Sebastian Riedel, Piotr
  Bojanowski, Armand Joulin, and Edouard Grave. 2022.
\newblock \href {https://openreview.net/forum?id=jKN1pXi7b0} {Unsupervised
  dense information retrieval with contrastive learning}.
\newblock \emph{Trans. Mach. Learn. Res.}, 2022.

\bibitem[{Izacard and Grave(2020)}]{izacard2020leveraging}
Gautier Izacard and Edouard Grave. 2020.
\newblock Leveraging passage retrieval with generative models for open domain
  question answering.
\newblock \emph{arXiv preprint arXiv:2007.01282}.

\bibitem[{Ji et~al.(2023)Ji, Lee, Frieske, Yu, Su, Xu, Ishii, Bang, Madotto,
  and Fung}]{DBLP:journals/csur/JiLFYSXIBMF23}
Ziwei Ji, Nayeon Lee, Rita Frieske, Tiezheng Yu, Dan Su, Yan Xu, Etsuko Ishii,
  Yejin Bang, Andrea Madotto, and Pascale Fung. 2023.
\newblock \href {https://doi.org/10.1145/3571730} {Survey of hallucination in
  natural language generation}.
\newblock \emph{{ACM} Comput. Surv.}, 55(12):248:1--248:38.

\bibitem[{Jiang et~al.(2023)Jiang, Xu, Gao, Sun, Liu, Dwivedi{-}Yu, Yang,
  Callan, and Neubig}]{active-rag}
Zhengbao Jiang, Frank~F. Xu, Luyu Gao, Zhiqing Sun, Qian Liu, Jane
  Dwivedi{-}Yu, Yiming Yang, Jamie Callan, and Graham Neubig. 2023.
\newblock \href {https://aclanthology.org/2023.emnlp-main.495} {Active
  retrieval augmented generation}.
\newblock In \emph{Proceedings of the 2023 Conference on Empirical Methods in
  Natural Language Processing, {EMNLP} 2023, Singapore, December 6-10, 2023},
  pages 7969--7992. Association for Computational Linguistics.

\bibitem[{Jin et~al.(2024)Jin, Zhang, Jiang, Liu, Liu, Liu, and
  Jin}]{jin2024ragcache}
Chao Jin, Zili Zhang, Xuanlin Jiang, Fangyue Liu, Xin Liu, Xuanzhe Liu, and Xin
  Jin. 2024.
\newblock Ragcache: Efficient knowledge caching for retrieval-augmented
  generation.
\newblock \emph{arXiv preprint arXiv:2404.12457}.

\bibitem[{Khandelwal et~al.(2019)Khandelwal, Levy, Jurafsky, Zettlemoyer, and
  Lewis}]{khandelwal2019generalization}
Urvashi Khandelwal, Omer Levy, Dan Jurafsky, Luke Zettlemoyer, and Mike Lewis.
  2019.
\newblock Generalization through memorization: Nearest neighbor language
  models.
\newblock \emph{arXiv preprint arXiv:1911.00172}.

\bibitem[{Leviathan et~al.(2023)Leviathan, Kalman, and
  Matias}]{leviathan2023fast}
Yaniv Leviathan, Matan Kalman, and Yossi Matias. 2023.
\newblock Fast inference from transformers via speculative decoding.
\newblock In \emph{International Conference on Machine Learning}, pages
  19274--19286. PMLR.

\bibitem[{Lewis et~al.(2020{\natexlab{a}})Lewis, Ghazvininejad, Ghosh,
  Aghajanyan, Wang, and Zettlemoyer}]{lewis2020pre}
Mike Lewis, Marjan Ghazvininejad, Gargi Ghosh, Armen Aghajanyan, Sida Wang, and
  Luke Zettlemoyer. 2020{\natexlab{a}}.
\newblock Pre-training via paraphrasing.
\newblock \emph{Advances in Neural Information Processing Systems},
  33:18470--18481.

\bibitem[{Lewis et~al.(2020{\natexlab{b}})Lewis, Perez, Piktus, Petroni,
  Karpukhin, Goyal, K{\"u}ttler, Lewis, Yih, Rockt{\"a}schel
  et~al.}]{lewis2020retrieval}
Patrick Lewis, Ethan Perez, Aleksandra Piktus, Fabio Petroni, Vladimir
  Karpukhin, Naman Goyal, Heinrich K{\"u}ttler, Mike Lewis, Wen-tau Yih, Tim
  Rockt{\"a}schel, et~al. 2020{\natexlab{b}}.
\newblock Retrieval-augmented generation for knowledge-intensive nlp tasks.
\newblock \emph{Advances in Neural Information Processing Systems},
  33:9459--9474.

\bibitem[{Lewis et~al.(2020{\natexlab{c}})Lewis, Perez, Piktus, Petroni,
  Karpukhin, Goyal, K{\"{u}}ttler, Lewis, Yih, Rockt{\"{a}}schel, Riedel, and
  Kiela}]{rag}
Patrick S.~H. Lewis, Ethan Perez, Aleksandra Piktus, Fabio Petroni, Vladimir
  Karpukhin, Naman Goyal, Heinrich K{\"{u}}ttler, Mike Lewis, Wen{-}tau Yih,
  Tim Rockt{\"{a}}schel, Sebastian Riedel, and Douwe Kiela. 2020{\natexlab{c}}.
\newblock \href
  {https://proceedings.neurips.cc/paper/2020/hash/6b493230205f780e1bc26945df7481e5-Abstract.html}
  {Retrieval-augmented generation for knowledge-intensive {NLP} tasks}.
\newblock In \emph{Advances in Neural Information Processing Systems 33: Annual
  Conference on Neural Information Processing Systems 2020, NeurIPS 2020,
  December 6-12, 2020, virtual}.

\bibitem[{Li et~al.(2022)Li, Su, Cai, Wang, and Liu}]{rag-survey}
Huayang Li, Yixuan Su, Deng Cai, Yan Wang, and Lemao Liu. 2022.
\newblock \href {http://arxiv.org/abs/2202.01110} {A survey on
  retrieval-augmented text generation}.
\newblock \emph{CoRR}, abs/2202.01110.

\bibitem[{Lin et~al.(2023)Lin, Tang, Tang, Yang, Dang, and Han}]{lin2023awq}
Ji~Lin, Jiaming Tang, Haotian Tang, Shang Yang, Xingyu Dang, and Song Han.
  2023.
\newblock Awq: Activation-aware weight quantization for llm compression and
  acceleration.
\newblock \emph{arXiv preprint arXiv:2306.00978}.

\bibitem[{Luo et~al.(2023)Luo, Zhang, Chuang, Gong, Kim, Wu, Meng, and
  Glass}]{sail}
Hongyin Luo, Tianhua Zhang, Yung{-}Sung Chuang, Yuan Gong, Yoon Kim, Xixin Wu,
  Helen Meng, and James~R. Glass. 2023.
\newblock \href {https://aclanthology.org/2023.findings-emnlp.242} {Search
  augmented instruction learning}.
\newblock In \emph{Findings of the Association for Computational Linguistics:
  {EMNLP} 2023, Singapore, December 6-10, 2023}, pages 3717--3729. Association
  for Computational Linguistics.

\bibitem[{Mallen et~al.(2023)Mallen, Asai, Zhong, Das, Khashabi, and
  Hajishirzi}]{popqa}
Alex Mallen, Akari Asai, Victor Zhong, Rajarshi Das, Daniel Khashabi, and
  Hannaneh Hajishirzi. 2023.
\newblock \href {https://doi.org/10.18653/V1/2023.ACL-LONG.546} {When not to
  trust language models: Investigating effectiveness of parametric and
  non-parametric memories}.
\newblock In \emph{Proceedings of the 61st Annual Meeting of the Association
  for Computational Linguistics (Volume 1: Long Papers), {ACL} 2023, Toronto,
  Canada, July 9-14, 2023}, pages 9802--9822. Association for Computational
  Linguistics.

\bibitem[{Min et~al.(2023)Min, Krishna, Lyu, Lewis, Yih, Koh, Iyyer,
  Zettlemoyer, and Hajishirzi}]{factscore}
Sewon Min, Kalpesh Krishna, Xinxi Lyu, Mike Lewis, Wen{-}tau Yih, Pang~Wei Koh,
  Mohit Iyyer, Luke Zettlemoyer, and Hannaneh Hajishirzi. 2023.
\newblock \href {https://aclanthology.org/2023.emnlp-main.741} {Factscore:
  Fine-grained atomic evaluation of factual precision in long form text
  generation}.
\newblock In \emph{Proceedings of the 2023 Conference on Empirical Methods in
  Natural Language Processing, {EMNLP} 2023, Singapore, December 6-10, 2023},
  pages 12076--12100. Association for Computational Linguistics.

\bibitem[{Muhlgay et~al.(2023)Muhlgay, Ram, Magar, Levine, Ratner, Belinkov,
  Abend, Leyton{-}Brown, Shashua, and Shoham}]{factuality-evaluation}
Dor Muhlgay, Ori Ram, Inbal Magar, Yoav Levine, Nir Ratner, Yonatan Belinkov,
  Omri Abend, Kevin Leyton{-}Brown, Amnon Shashua, and Yoav Shoham. 2023.
\newblock \href {https://doi.org/10.48550/ARXIV.2307.06908} {Generating
  benchmarks for factuality evaluation of language models}.
\newblock \emph{CoRR}, abs/2307.06908.

\bibitem[{Ouyang et~al.(2022)Ouyang, Wu, Jiang, Almeida, Wainwright, Mishkin,
  Zhang, Agarwal, Slama, Ray, Schulman, Hilton, Kelton, Miller, Simens, Askell,
  Welinder, Christiano, Leike, and Lowe}]{gpt3.5}
Long Ouyang, Jeffrey Wu, Xu~Jiang, Diogo Almeida, Carroll~L. Wainwright, Pamela
  Mishkin, Chong Zhang, Sandhini Agarwal, Katarina Slama, Alex Ray, John
  Schulman, Jacob Hilton, Fraser Kelton, Luke Miller, Maddie Simens, Amanda
  Askell, Peter Welinder, Paul~F. Christiano, Jan Leike, and Ryan Lowe. 2022.
\newblock \href
  {http://papers.nips.cc/paper\_files/paper/2022/hash/b1efde53be364a73914f58805a001731-Abstract-Conference.html}
  {Training language models to follow instructions with human feedback}.
\newblock In \emph{NeurIPS}.

\bibitem[{Ratner et~al.(2022)Ratner, Levine, Belinkov, Ram, Magar, Abend,
  Karpas, Shashua, Leyton-Brown, and Shoham}]{ratner2022parallel}
Nir Ratner, Yoav Levine, Yonatan Belinkov, Ori Ram, Inbal Magar, Omri Abend,
  Ehud Karpas, Amnon Shashua, Kevin Leyton-Brown, and Yoav Shoham. 2022.
\newblock Parallel context windows for large language models.
\newblock \emph{arXiv preprint arXiv:2212.10947}.

\bibitem[{Reid et~al.(2024)Reid, Savinov, Teplyashin, Lepikhin, Lillicrap,
  Alayrac, Soricut, Lazaridou, Firat, Schrittwieser et~al.}]{reid2024gemini}
Machel Reid, Nikolay Savinov, Denis Teplyashin, Dmitry Lepikhin, Timothy
  Lillicrap, Jean-baptiste Alayrac, Radu Soricut, Angeliki Lazaridou, Orhan
  Firat, Julian Schrittwieser, et~al. 2024.
\newblock Gemini 1.5: Unlocking multimodal understanding across millions of
  tokens of context.
\newblock \emph{arXiv preprint arXiv:2403.05530}.

\bibitem[{Schick et~al.(2023)Schick, Dwivedi{-}Yu, Dess{\`{\i}}, Raileanu,
  Lomeli, Zettlemoyer, Cancedda, and Scialom}]{toolformer}
Timo Schick, Jane Dwivedi{-}Yu, Roberto Dess{\`{\i}}, Roberta Raileanu, Maria
  Lomeli, Luke Zettlemoyer, Nicola Cancedda, and Thomas Scialom. 2023.
\newblock \href {https://doi.org/10.48550/ARXIV.2302.04761} {Toolformer:
  Language models can teach themselves to use tools}.
\newblock \emph{CoRR}, abs/2302.04761.

\bibitem[{Shazeer(2019)}]{shazeer2019fast}
Noam Shazeer. 2019.
\newblock Fast transformer decoding: One write-head is all you need.
\newblock \emph{arXiv preprint arXiv:1911.02150}.

\bibitem[{Shazeer and Stern(2018)}]{shazeer2018adafactor}
Noam Shazeer and Mitchell Stern. 2018.
\newblock Adafactor: Adaptive learning rates with sublinear memory cost.
\newblock In \emph{International Conference on Machine Learning}, pages
  4596--4604. PMLR.

\bibitem[{Shi et~al.(2023)Shi, Chen, Misra, Scales, Dohan, Chi, Sch\"{a}rli,
  and Zhou}]{irrelevant-ctx}
Freda Shi, Xinyun Chen, Kanishka Misra, Nathan Scales, David Dohan, Ed~H. Chi,
  Nathanael Sch\"{a}rli, and Denny Zhou. 2023.
\newblock \href {https://proceedings.mlr.press/v202/shi23a.html} {Large
  language models can be easily distracted by irrelevant context}.
\newblock In \emph{Proceedings of the 40th International Conference on Machine
  Learning}, volume 202 of \emph{Proceedings of Machine Learning Research},
  pages 31210--31227. PMLR.

\bibitem[{Team et~al.(2023)Team, Anil, Borgeaud, Wu, Alayrac, Yu, Soricut,
  Schalkwyk, Dai, Hauth et~al.}]{team2023gemini}
Gemini Team, Rohan Anil, Sebastian Borgeaud, Yonghui Wu, Jean-Baptiste Alayrac,
  Jiahui Yu, Radu Soricut, Johan Schalkwyk, Andrew~M Dai, Anja Hauth, et~al.
  2023.
\newblock Gemini: a family of highly capable multimodal models.
\newblock \emph{arXiv preprint arXiv:2312.11805}.

\bibitem[{Touvron et~al.(2023)Touvron, Lavril, Izacard, Martinet, Lachaux,
  Lacroix, Rozi{\`{e}}re, Goyal, Hambro, Azhar, Rodriguez, Joulin, Grave, and
  Lample}]{llama}
Hugo Touvron, Thibaut Lavril, Gautier Izacard, Xavier Martinet, Marie{-}Anne
  Lachaux, Timoth{\'{e}}e Lacroix, Baptiste Rozi{\`{e}}re, Naman Goyal, Eric
  Hambro, Faisal Azhar, Aur{\'{e}}lien Rodriguez, Armand Joulin, Edouard Grave,
  and Guillaume Lample. 2023.
\newblock \href {https://doi.org/10.48550/ARXIV.2302.13971} {Llama: Open and
  efficient foundation language models}.
\newblock \emph{CoRR}, abs/2302.13971.

\bibitem[{Yan et~al.(2024)Yan, Gu, Zhu, and Ling}]{yan2024corrective}
Shi-Qi Yan, Jia-Chen Gu, Yun Zhu, and Zhen-Hua Ling. 2024.
\newblock Corrective retrieval augmented generation.
\newblock \emph{arXiv preprint arXiv:2401.15884}.

\bibitem[{Yogatama et~al.(2021)Yogatama, de~Masson~d’Autume, and
  Kong}]{yogatama2021adaptive}
Dani Yogatama, Cyprien de~Masson~d’Autume, and Lingpeng Kong. 2021.
\newblock Adaptive semiparametric language models.
\newblock \emph{Transactions of the Association for Computational Linguistics},
  9:362--373.

\bibitem[{Yoran et~al.(2023)Yoran, Wolfson, Ram, and Berant}]{ralm-nli}
Ori Yoran, Tomer Wolfson, Ori Ram, and Jonathan Berant. 2023.
\newblock \href {https://doi.org/10.48550/ARXIV.2310.01558} {Making
  retrieval-augmented language models robust to irrelevant context}.
\newblock \emph{CoRR}, abs/2310.01558.

\bibitem[{Zhang et~al.(2023)Zhang, Li, Cui, Cai, Liu, Fu, Huang, Zhao, Zhang,
  Chen, Wang, Luu, Bi, Shi, and Shi}]{hallucination-survey}
Yue Zhang, Yafu Li, Leyang Cui, Deng Cai, Lemao Liu, Tingchen Fu, Xinting
  Huang, Enbo Zhao, Yu~Zhang, Yulong Chen, Longyue Wang, Anh~Tuan Luu, Wei Bi,
  Freda Shi, and Shuming Shi. 2023.
\newblock \href {https://doi.org/10.48550/ARXIV.2309.01219} {Siren's song in
  the {AI} ocean: {A} survey on hallucination in large language models}.
\newblock \emph{CoRR}, abs/2309.01219.

\bibitem[{Zhong et~al.(2021)Zhong, Yin, Yu, Zaidi, Mutuma, Jha, Awadallah,
  Celikyilmaz, Liu, Qiu et~al.}]{zhong2021qmsum}
Ming Zhong, Da~Yin, Tao Yu, Ahmad Zaidi, Mutethia Mutuma, Rahul Jha,
  Ahmed~Hassan Awadallah, Asli Celikyilmaz, Yang Liu, Xipeng Qiu, et~al. 2021.
\newblock Qmsum: A new benchmark for query-based multi-domain meeting
  summarization.
\newblock \emph{arXiv preprint arXiv:2104.05938}.

\end{thebibliography}
\bibliographystyle{acl_natbib}

\newpage
\appendix

\section{Prompts and Instructions}

\subsection{Prompts Used for LLMs}

We share the prompt used for calling LLMs to get per context assessment in Table~\ref{tab:critique_prompt}. 
\begin{table*}[h]
\centering
\caption{The zero-shot prompts for LLM labeling and critique.}
\fontsize{9pt}{9pt}\selectfont
\setlength{\tabcolsep}{1.2pt}
\begin{tabular}{p{0.95\linewidth}}
\toprule
\hspace{0.5cm}Round 1 prompt\\ 
\midrule

\hspace{0.5cm}\begin{tabular}[c]{@{}p{0.9\linewidth}@{}}
You are now doing a reading comprehension task. It is important that you be as thorough, detail-oriented, and accurate as possible in your response. \\

You are given a question, a set of accepted answers, a document and its title. The document does not necessarily contain the right answer to the question. \\

You should read the title and the document and then check if they provide one  of the correct answers to the question. \\

If the title and document together contain the correct answer to the question, output a score of 1.0, otherwise output a score of 0.0. \\

   question: <question> \\
   accepted answers: <answers> \\
   title: <title> \\
   document: <document> \\
   output:\\
\end{tabular}\\
\midrule
\hspace{0.5cm}Round 2 prompt\\ 
\midrule
\hspace{0.5cm}\begin{tabular}[c]{@{}p{0.9\linewidth}@{}}
Your job is to correct another model's performance on a reading comprehension task. \\

The model was given a question, a set of accepted answers, a document and its title. The document and title do not necessarily contain the right answer. The model was instructed to output a score of 1.0 if the document contains the answer, and a score of 0.0 otherwise. \\

You will be given the same information as the other model along with its output. You should read the title and document and then check if they provide one of the correct answers to the question. \\

Then check if you agree with the previous model's output. \\
If you agree, output the same score unchanged. \\
If you disagree, output the corrected score. \\
Your output should be as accurate as possible. \\

   question: <question> \\
   accepted answers: <answers> \\
   title: <title> \\
   document: <document> \\
   previous model's score: <score> \\
   output:
\end{tabular}\\

\bottomrule

\end{tabular}
\label{tab:critique_prompt}
\end{table*}

\subsection{Rater Guidelines}
\label{rater}

We share the instructions provided to the human labelers in Table~\ref{tab:rater}.

\begin{table*}[h]
\centering
\caption{Instructions for raters creating ground-truth relevance dataset.}
\fontsize{9pt}{9pt}\selectfont
\setlength{\tabcolsep}{1.2pt}
\begin{tabular}{p{0.95\linewidth}}
\toprule
\hspace{0.5cm}Human Rater Instructions\\ 
\midrule

\hspace{0.5cm}\begin{tabular}[c]{@{}p{0.9\linewidth}@{}}
Please read the question, the answer and the context. Please answer if the context can help answer the question. If it can, select 1. Otherwise select 0. \\
\\
1: good \\
0: bad \\
\\
Please use the answers as a hint. However, do not use "is the answer in the context?" as a heuristic for making the decision. \\
\end{tabular}\\
\bottomrule

\end{tabular}
\label{tab:rater}
\end{table*}

\section{Dataset Analysis}
\label{data_concerns}

During the human-labeling process, several raters flagged documents and questions that were difficult to label. In total, 23 out of 500 documents were flagged and 15 out of 50 questions were flagged.

We observed trends in questions and contexts shared in Table~\ref{tab:concerns_table} that raised concerns about whether and how a human would be able to assess the relevance of the context. These concerns extend to expectations of how well LLMs would be able to do that as well.

\begin{table}[!h]
\centering
\caption{Overview of trends, datasets, and examples with associated comments.}
\setlength{\arrayrulewidth}{0.3mm} 
\setlength{\tabcolsep}{8pt} 
\renewcommand{\arraystretch}{1.5} 
\resizebox{0.98\linewidth}{!}{
\begin{tabular}{|m{2.0cm}|m{1.0cm}|m{3.0cm}|m{6cm}|}
\hline
\rowcolor[gray]{0.9} \textbf{Trend} & \textbf{Dataset} & \textbf{Example question} & \textbf{Comments} \\ \hline
\multirow{6}{=}{Questions with time-dependent answers} & NQ & who is the president of usa right now & Depends on when the question is asked. \\ \cline{2-4}
 & NQ & who is the current director of the us mint & Depends on when the question is asked. \\ \cline{2-4}
 & NQ & when is the next deadpool movie being released & Depends on when the question is asked. \\ \cline{2-4}
 & NQ & total number of death row inmates in the us & Fluctuates over time. \\ \cline{2-4}
 & PopQA & What is Prague the capital of? & The borders in this region and the name of the country have changed several times in the 20th century. \\ \cline{2-4}
 & PopQA & What is Dennis Rodman's occupation? & The accepted answers are "actor, actress, actors, actresses". He was an actor later in his career, but he rose to prominence as a professional basketball player. \\ \hline
\multirow{2}{=}{Missing synonyms, abbreviations, and aliases, combined with unclear granularity} & NQ & in which regions are most of Africa petroleum and natural gas found & "Region" can refer to different levels of granularity (e.g. Sub-Saharan Africa vs. Ethiopia), but the only accepted answer is "Nigeria". \\ \cline{2-4}
 & NQ & what type of car is a jeep & The accepted answers are only "off-road vehicles", "light utility vehicles", "sport utility vehicles", but "SUV" is clearly a correct answer as well. \\ \hline
Non-exhaustive list of answers & NQ & cast of law \& order special victim unit & The accepted answers include 16 cast members, but the show went on for 25 seasons with many cast changes and guest stars not included in the list. \\ \hline
Oddly phrased question & NQ & right to property according to the constitution of India is a & The only correct answer is "constitutional right", but that is included in the question. It's not clear what type of answer would be appropriate here. \\ \hline
Overly specific answers expected & NQ & where does the story the great gatsby take place & The only accepted answer here is "Long Island of 1922", but the place is Long Island and the question does not ask about when the story is set. \\ \hline
\multirow{3}{=}{Question refers to an entity with a common name without disambiguation} & PopQA & What genre is Frances? & There is a musician and a film called "Frances", and both of those could arguably have a genre associated with them. \\ \cline{2-4}
 & PopQA & Who was the producer of Hurt? & The question is referring to a song performed by Christina Aguilera but there are many other songs, movies, and other entities that share the name and also have a producer. \\ \cline{2-4}
 & PopQA & What is the capital of Cherokee County? & There are many different Cherokee Counties in different states in the USA. \\ \hline
\end{tabular}
}
\label{tab:concerns_table}
\end{table}

We explored several ways of filtering our human-labeled subsample of Natural Questions to determine how they impacted context assessment F-scores overall and for each auto-rater. We provide two additional filtered versions of the human-labeled RAG relevance dataset as alternatives. See Table~\ref{tab:ground_truth_filtering_stats} for statistics on the filtered datasets and Table~\ref{tab:labeling-methods} for the auto-rater F-scores for each. Both statistical filtering approaches (e.g. removing contexts with non-unanimous labels) and targeted filtering approaches (e.g. removing questions or contexts flagged by human raters) lead to some improvement in F-scores for relevance labels.

\begin{table}[ht]
\centering
\caption{Summary of Dataset Evaluations}
\resizebox{0.98\linewidth}{!}{
\begin{tabular}{l c c c c c c c}
\toprule
\textbf{Dataset} & \textbf{\# Questions} & \textbf{\# Docs} & \textbf{Total \% Relevant} & \textbf{Average \# Docs per Question} & \textbf{Std Dev \# Docs per Question} & \textbf{Average \% relevant docs per question} & \textbf{Std Dev \% relevant docs per question} \\
\midrule
Original 3-rater labels & 50 & 500 & 0.30 & 10 & 0 & 0.30 & 0.23 \\
Revisited (released) & 50 & 500 & 0.31 & 10 & 0 & 0.31 & 0.26 \\
Filter non-unanimous (released) & 50 & 351 & 0.23 & 7.02 & 2.59 & 0.27 & 0.29 \\
Filter flagged docs & 48 & 467 & 0.31 & 9.729 & 0.57 & 0.31 & 0.27 \\
Filter flagged questions & 35 & 350 & 0.29 & 10 & 0 & 0.29 & 0.26 \\
Filter flagged docs and question (released) & 34 & 330 & 0.29 & 9.7 & 0.58 & 0.29 & 0.27 \\
Filter 4-rater split & 50 & 444 & 0.28 & 8.88 & 1.49 & 0.29 & 0.27 \\
\bottomrule
\end{tabular}
}
\label{tab:ground_truth_filtering_stats}
\end{table}

\begin{table}[!ht]
\centering
\caption{Evaluation of Labeling Methods w/ Filtering}
\label{tab:labeling-methods}
\resizebox{0.98\linewidth}{!}{
\begin{tabular}{p{3.0cm} p{1.5cm} c c c} 
\hline
Dataset Filters & \# Docs &  Rater Model & Critic Model & Average F1 \\
\hline
\multirow{6}{=}{Specialized rater corrections} & \multirow{6}{=}{500} & PALM2 XL & n/a & 0.729 \\
 & & PALM2 XL & PALM2 XL & 0.781 \\
 & & Gemini Ultra & n/a & 0.761 \\
 & & Gemini Ultra & Gemini Ultra & 0.704 \\
 & & PALM2 XL & Gemini Ultra & 0.728 \\
 & & Gemini Ultra & PALM2 XL 340B & \textbf{0.821} \\
\hline
\multirow{6}{=}{Filter non-unanimous docs} & \multirow{6}{=}{351} & PALM2 XL & n/a/ & 0.741 \\
 & & PALM2 XL & PALM2 XL & 0.811 \\
 & & Gemini Ultra & n/a & 0.792 \\
 & & Gemini Ultra & Gemini Ultra & 0.739 \\
 & & PALM2 XL & Gemini Ultra & 0.763 \\
 & & Gemini Ultra & PALM2 XL & \textbf{0.856} \\
\hline
\multirow{6}{=}{Filter flagged docs and questions} & \multirow{6}{=}{330} & PALM2 XL & n/a & 0.750 \\
 & & PALM2 XL & PALM2 XL & 0.797 \\
 & & Gemini Ultra & n/a & 0.782 \\
 & & Gemini Ultra & Gemini Ultra & 0.741 \\
 & & PALM2 XL & Gemini Ultra & 0.753 \\
 & & Gemini Ultra & PALM2 XL & \textbf{0.833} \\
\hline
\end{tabular}
}
\end{table}

\end{document}